\documentclass[10pt,twocolumn,letterpaper]{article}
\pdfoutput=1
\usepackage{iccv}
\usepackage[accsupp]{axessibility}
\usepackage{times}
\usepackage{epsfig}
\usepackage{graphicx}
\usepackage{amsmath}
\usepackage{amssymb}
\usepackage{multirow}
\usepackage{xcolor}
\usepackage{subfig}
\usepackage{booktabs}
\usepackage{colortbl}
\usepackage[normalem]{ulem}
\useunder{\uline}{\ul}{}

\definecolor{mygray}{gray}{0.926}
\definecolor{LightCyan}{rgb}{0.935,1,1}
\usepackage{hyperref}
\hypersetup{pagebackref=true,breaklinks=true,letterpaper=true,colorlinks,bookmarks=false}

\iccvfinalcopy % *** Uncomment this line for the final submission

\def\iccvPaperID{8889} % *** Enter the ICCV Paper ID here

\ificcvfinal\fi

\begin{document}

%%%%%%%%% TITLE
\title{WaveNeRF: Wavelet-based Generalizable Neural Radiance Fields}

\author{Muyu Xu$^{1}$
\quad Fangneng Zhan$^{2}$
\quad Jiahui Zhang$^{1}$
\quad Yingchen Yu$^{1}$\\
\quad Xiaoqin Zhang$^{3}$
\quad Christian Theobalt$^{2}$
\quad Ling Shao$^{4}$
\quad Shijian Lu$^{1}$\\
\\[1mm]
{\small $^1$Nanyang Technological University\quad$^2$Max Planck Institute for Informatics}\\[0.1mm]
{\small $^3$Wenzhou University\quad$^4$UCAS-Terminus AI Lab, UCAS}
\vspace{-1em}
% To save space, use either the email address or home page.
}

\maketitle
\ificcvfinal\thispagestyle{empty}\fi

\def\thefootnote{*}\footnotetext{Shijian Lu is the corresponding author.}

\begin{figure*}[htb]
\centering
\subfloat[Ground Truth (GT)]{
\includegraphics[width=0.25\textwidth]{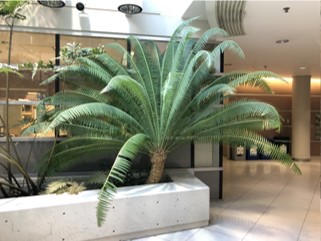}
}
\subfloat[Rendered Result]{
\includegraphics[width=0.25\textwidth]{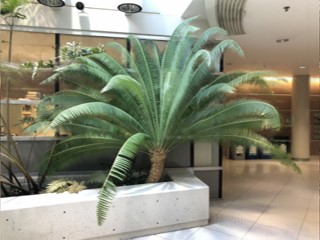}
}
\subfloat[Absolute Rendering Errors]{
\includegraphics[width=0.25\textwidth]{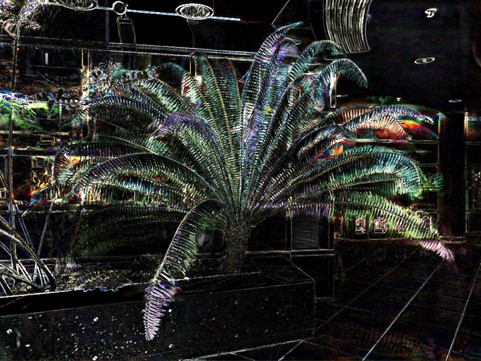}
}
\subfloat[High-Frequency Features of GT]{
\includegraphics[width=0.25\textwidth]{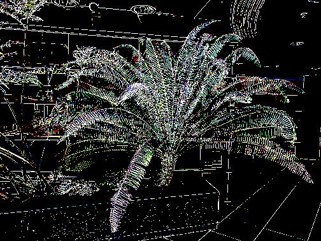}
}
\caption{The comparison between the absolute rendering errors (c) of GeoNeRF\cite{johari2022geonerf} and the high-frequency features of the ground truth (d). We can see that the errors mainly appear around the pixels with high-frequency features.}
\label{fig:obs}
\end{figure*}

%%%%%%%%% ABSTRACT
\begin{abstract}
Neural Radiance Field (NeRF) has shown impressive performance in novel view synthesis via implicit scene representation.
However, it usually suffers from poor scalability as requiring densely sampled images for each new scene. Several studies have attempted to mitigate this problem by integrating Multi-View Stereo (MVS) technique into NeRF while they still entail a cumbersome fine-tuning process for new scenes.
Notably, the rendering quality will drop severely without this fine-tuning process and the errors mainly appear around the high-frequency features.
In the light of this observation, we design WaveNeRF, which integrates wavelet frequency decomposition into MVS and NeRF to achieve generalizable yet high-quality synthesis without any per-scene optimization.
To preserve high-frequency information when generating 3D feature volumes, WaveNeRF builds Multi-View Stereo in the Wavelet domain by integrating the discrete wavelet transform into the classical cascade MVS, which disentangles high-frequency information explicitly.
With that, disentangled frequency features can be injected into classic NeRF via a novel hybrid neural renderer to yield faithful high-frequency details, and an intuitive frequency-guided sampling strategy can be designed to suppress artifacts around high-frequency regions.
Extensive experiments over three widely studied benchmarks show that WaveNeRF achieves superior generalizable radiance field modeling when only given three images as input.

\end{abstract}

% %%%%%%%%% BODY TEXT
\section{Introduction}
Rendering novel views from a set of posed scene images has been studied for years in the fields of computer vision and graphics. With the emergence of implicit neural representation, neural radiance field (NeRF)\cite{mildenhall2021nerf} and its variants\cite{liu2020neural, martin2021nerf} have recently achieved very impressive performance in novel view synthesis with superb multi-view consistency. However, most existing works fall short in model scalability by requiring a per-scene optimization process with densely sampled multi-view images for training.
% a fair number of posed training images that are densely sampled from multiple viewpoints.

To avoid the cumbersome process of training from scratch for new scenes,
a popular line of generalizable NeRF~\cite{chen2021mvsnerf, yu2021pixelnerf, wang2021ibrnet, xu2022point, johari2022geonerf} introduces a pipeline that first trains a base model on the training data and then conducts fine-tuning for each new scene, which improves the scalability and shortens the per-scene training process. Their base models often extract features from the source views and then inject the features into a neural radiance field. Several previous studies~\cite{yu2021pixelnerf, wang2021ibrnet} directly use CNN networks to extract features while recent generalizable NeRF models~\cite{chen2021mvsnerf, xu2022point, johari2022geonerf} resort to Multi-View Stereo (MVS) technique to warp 2D source feature maps into 3D features planes, yielding better performance than merely using CNN networks.
% With , generalizable NeRF~\cite{chen2021mvsnerf, yu2021pixelnerf, wang2021ibrnet, xu2022point, johari2022geonerf} 
However, per-scene fine-tuning still entails a fair number of posed training images that are often challenging to collect in various real-world tasks. On the other hand, removing the per-scene fine-tuning will incur a significant performance drop with undesired artifacts and poor detail.
% without the per-scene fine-tuning process. 
Notably, we intriguingly observe that the rendering error mainly lies around image regions with rich high-frequency information as illustrated in Fig.~\ref{fig:obs}.
% by comparing the rendered error of the models without fine-tuning with the high-frequency feature maps of the ground truth (see Figure \ref{fig:obs}),
The phenomenon of losing high-frequency detail is largely attributed to the fact that most existing generalizable NeRFs conduct down-sampling operations at the feature extraction stage of their pipeline, i.e., the CNN networks adopted in ~\cite{yu2021pixelnerf, wang2021ibrnet} or the MVS module adopted in ~\cite{chen2021mvsnerf, xu2022point, johari2022geonerf}.

In the light of the aforementioned observation, 
we present \textbf{Wave}lets-based \textbf{Ne}ural \textbf{R}adiance \textbf{F}ields (\textbf{WaveNeRF}) which incorporates explicit high-frequency information into the training process and thus obviates the per-scene fine-tuning under the generalizable and few-shot setting.
% The proposed WaveNeRF is capable of synthesizing photorealistic novel views of new scenes using just three-shot posed training images.
% without any per-scene fine-tuning. 
% While many recent studies on generalizable NeRF utilize multi-view stereo (MVS) techniques to construct 3D feature volumes from source views and then convert them into NeRF, 
% Specifically, WaveNeRF employs multi-view stereo (MVS) technique to construct 3D feature volumes from source views and then convert them to model radiance fields, where the wavelet coefficients
Specifically, with MVS technique to construct 3D feature volumes which are converted to model NeRF in the spatial domain,
we further design a Wavelet Multi-View Stereo (WMVS) to incorporate scene wavelet coefficients into the MVS to achieve frequency-domain modeling.
Distinct from other frequency transformations like Fourier Transform, WaveNeRF employs Wavelet Transform which is coordinate invariant and preserves the relative spatial positions of pixels. 
This property is particularly advantageous in the context of MVS as it allows multiple input views to be warped in the direction of a reference view to form sweeping planes in both the spatial domain and the frequency domain within the same coordinate system. 
% Moreover, this property also allows 
Apart from MVS, this property also enables to build a frequency-based radiance field so that a designed Hybrid Neural Renderer (HNR) can leverage the information in both the spatial and frequency domains to boost the rendering quality of the appearance, especially around the high-frequency regions.
In addition, WaveNeRF is also equipped with a Frequency-guided Sampling Strategy (FSS) which enables the model to focus on the regions with larger high-frequency coefficients.
% leverages frequency features to guide the sampling process in NeRF,  
% Thus, 
The rendering quality can be improved clearly with FSS by sampling denser points around object surfaces.
% FSS not only benefits the rendering quality in locations where previous studies have larger errors but also leads to an overall better rendering quality by 

The contributions of this work can be summarized in three points. 
\begin{itemize}
    \item \textit{First}, we design a WMVS module that preserves high-frequency information effectively by incorporating wavelet frequency volumes while extracting geometric scene features.
    \item \textit{Second}, we design a HNR module that can merge the features from both the spatial domain and the frequency domain, yielding faithful high-frequency details in neural rendering. 
    \item \textit{Third}, we develop FSS that can guide the volume rendering to sample denser points around the object surfaces so that it can infer the appearance and geometry with higher quality.
\end{itemize}

\section{Related Works}
\subsection{Multi-View Stereo}
Multi-view stereo (MVS) is a method that involves using multiple images taken from various viewpoints to create a detailed 3D reconstruction of an object or scene. Over time, various conventional methods have been proposed and tested in this field ~\cite{de1999poxels, kolmogorov2002multi, esteban2004silhouette, seitz2006comparison, furukawa2009accurate, schonberger2016pixelwise}. More recently, deep learning techniques have been integrated into the multi-view stereo process. One such technique is MVSNet~\cite{yao2018mvsnet}, which extracts features from all input images and warps them onto a reference image to generate probabilistic planes with varying depth values. These planes are then combined to create a variance-based cost volume that accurately represents the specific scene.

Although MVS methods have demonstrated promising performance, their large memory requirements, due to the 3D volume grid and operations, severely limit the resolution of input images and subsequent development of deep learning-based MVS research. To address this issue, R-MVSNet~\cite{yao2019recurrent} sequentially regularizes the cost volume with GRU, making MVSNet more scalable. 
% Although the performance of R-MVSNet is slightly worse than the state-of-the-art, it offers an approach to reduce memory costs and enables deployment on larger-scale scenes. 
In addition, cascade MVS models~\cite{cheng2020deep, gu2020cascade} use a coarse-to-fine strategy to generate cost volumes of various scales and compute depth output accordingly, freeing up more memory space. 
MVS has been shown to be effective in inferring the geometry and occlusions of a scene~\cite{chen2021mvsnerf, johari2022geonerf}. We follow the previous MVS techniques and further introduce wavelet transform into it to achieve a higher quality of inference.

\subsection{Neural Radiance Field}
3D scene reconstruction and novel view synthesis have been extensively studied for many years. Researchers have used various explicit representations of scene geometry such as 3D meshes~\cite{kazhdan2006poisson, lorensen1987marching} and point clouds~\cite{lassner2021pulsar, aliev2020neural}. However, NeRF~\cite{mildenhall2021nerf} employs an implicit neural representation method that uses an MLP-based network to render novel views. NeRF has demonstrated excellent rendering performance and has been further extended to various computer vision tasks~\cite{khalid2022wildnerf, chen2021hallucinated, pumarola2021d, gafni2021dynamic, li2021neural, gao2021dynamic, schwarz2020graf, chan2021pi, sucar2021imap, zhang2022vmrf, liu2023stylerf, liu20233d}. Although all of these studies showcase the impressive strength of NeRF in specific tasks, they still follow the same training process as the original NeRF and require per-scene training to complete the corresponding task.

To address this issue, several studies in the generalization of NeRF have shown some degree of success. Specifically, PixelNeRF~\cite{yu2021pixelnerf} and IBRNet~\cite{wang2021ibrnet} both rely on the notion that aggregating multi-view features at each sampled point leads to better performance than using direct encoded RGB inputs. Another typical approach that achieves generalizable NeRF is using multi-view stereo (MVS) techniques. For instance, MVSNeRF~\cite{chen2021mvsnerf}, which is the first to combine MVSNet and NeRF, simply concatenates the cost volume in MVSNet with the 5D input in NeRF. More recent generalizable NeRF, PointNeRF~\cite{xu2022point} and GeoNeRF~\cite{johari2022geonerf}, both use MVS techniques to obtain a coarse 3D representation, but PointNeRF uses point cloud growing to enhance the inference ability, while GeoNeRF uses attention-based transformer modules.

Although some of the NeRF models are generalizable, they typically require a specific number of inputs, such as 10 source views in IBRNet~\cite{wang2021ibrnet}. In addition, almost all of them need per-scene optimization to achieve photorealistic outcomes. Per-scene optimization is actually an additional training process which greatly impairs the generalizability. It is worth noting that without this optimization process, the rendering quality of these existing models can drop significantly, with most errors occurring around high-frequency features. Based on this observation, we integrate wavelet frequency decomposition into NeRF to achieve generalizable yet high-quality synthesis without any per-scene optimization. We believe that this approach is much more realistic, as it mimics situations where intelligent vehicles have limited sensors and need to reconstruct 3D scenery immediately.

\section{Method}
This section presents our novel wavelet-based generalizable NeRF, designed for synthesizing high-quality novel views of a scene from three-shot source views without any per-scene fine-tuning process.
Inspired by the observation that the rendering errors of the previous models mainly gather around the high-frequency regions, we design a Wavelet Multi-view Stereo (WMVS) module to obtain feature volumes in both the spatial domain and the frequency domain so that the high-frequency information can be maintained and represented separately. Besides, since the renderer in prior studies is unable to directly disentangle the errors around high-frequency features, we implement a Hybrid Neural Renderer (HNR) that can adjust the rendered colors based on the high-frequency information obtained from WMVS. 
During this rendering process, we also notice that previous sampling strategies necessitate an additional sampling step based on the outcome of the initial sampling, or they simultaneously sample all the points at the expense of sampling quality.
Therefore, to achieve higher sampling quality where more samples are around objects in the scene in a one-round sampling process, we adopt a Frequency-guided Sampling Strategy (FSS) where the %fine sampling process is guided by the
 coordinates of the sampled points are determined by the distribution of the features in the frequency feature volume.
% Each sample contains a view-independent feature token $t_{n, 0}$ and V view-dependent feature tokens $\{t_{n, v}\}^{N,V}_{n=0,v=0}$ in both the spatial domain and the frequency domain. These tokens are then fed into a subsequent Hybrid Neural Renderer (HNR) which is used to infer the volume density, colors, and frequency coefficients at the sample locations based on these tokens (see Section \ref{sec3.3}). At last, the final colors and frequency coefficients are inferred via classical volume rendering.

\begin{figure*}[htb]
    \centering
    \includegraphics[width=1.\textwidth]{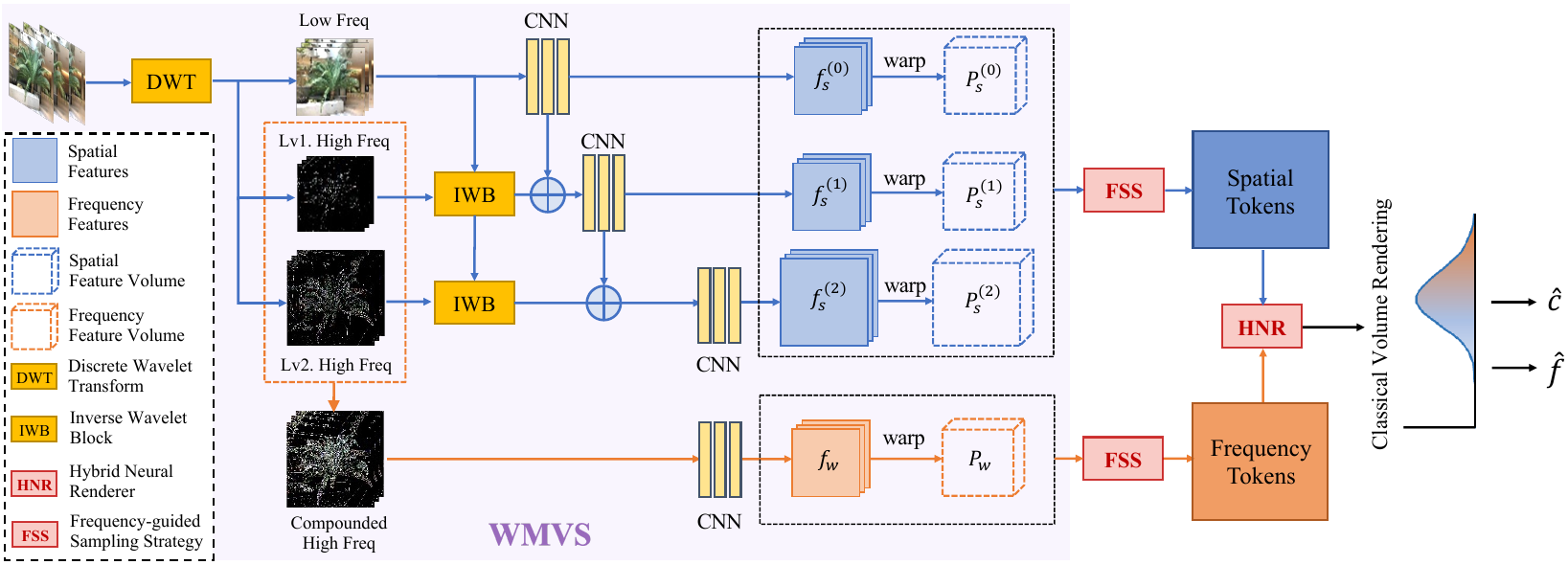}
    \caption{%The overview of WaveNeRF. Our pipeline first utilizes a Wavelet Multi-view Stereo (WMVS) module to obtain feature volumes in both the spatial space and the frequency space. Thereafter, to obtain more precise samples around objects in the scene, we adopt a special course-to-fine sampling strategy where the fine sampling process is guided by the distribution of the coarse samples in the frequency feature volume. Each sample contains a view-independent feature token $t_{n, 0}$ and V view-dependent feature tokens $\{t_{n, v}\}^{N,V}_{n=0,v=0}$ in both the spatial domain and the frequency domain. These tokens are then fed into a subsequent Hybrid Neural Renderer (HNR) which is used to infer the volume density, colors, and frequency coefficients at the sample locations based on these tokens (see Section \ref{sec3.3}). At last, the final colors and frequency coefficients are inferred via classical volume rendering.
    The overview of the proposed WaveNeRF. With sparse input views, wavelet multi-view stereo (WMVS) is designed to produce frequency feature volume $f_w$ and multi-level spatial feature volumes $[f^{(0)}_s, f^{(1)}_s, f^{(2)}_s]$. Specifically, the input views are first divided into different frequency components with level-2 discrete wavelet transform. The spatial and frequency features are then obtained via our designed Inverse Wavelet Blocks (IWB) and CNN-based feature extractors, and warpped into corresponding 3D feature volumes$ [P^{(0)}_s, P^{(1)}_s, P^{(2)}_s, P_w]$. With 2D features and 3D volumes, a novel Frequency-guided Sampling Strategy (FSS) is introduced to yield more precise samples with spatial and frequency tokens. These tokens are fed into a subsequent Hybrid Neural Renderer (HNR) to infer the volume density, colors, and frequency coefficients.}
    \label{fig:model}
\end{figure*}

The overall architecture of WaveNeRF is shown in Fig. \ref{fig:model}. We elaborate on our designed WMVS, FSS, and HNR, in Section \ref{sec3.1}, \ref{sec3.2}, and \ref{sec3.3} respectively.

\subsection{Wavelet Multi-view Stereo}
\label{sec3.1}
Since Wavelet Transform can decompose an image into components with different scales, it naturally fits with the pyramid structure of the CasMVSNet\cite{gu2020cascade}. Therefore, given a set of input source views $\{I_v\}^V_{v=0}$ with the size of $H \times W$, we design a Wavelet Multi-View Stereo (WMVS) module to construct cascaded spatial feature volumes as well as a high-frequency feature volume following the similar way of CasMVSNet as shown in Fig. \ref{fig:model}. We make several modifications to both the feature extraction process and the volume construction process of CasMVSNet. First, 
% we design a Wavelet Feature Pyramid Network (WFPN) to generate 2D semantic feature maps and a 2D wavelet feature map, which can avoid the information loss caused by the direct downsampling operation in CasMVSNet as below:
% \begin{align}
%     f_s^{(l)}, f_w^{(l)} =& \textbf{WFPN}(I_v) \in \mathbf{R}^{\frac{H}{2^l}\times\frac{W}{2^l}}, ~~l\in{0, 1, 2} 
% \end{align}
% \noindent where $f_s$ and $f_w$ indicate the spatial feature maps and wavelet feature maps respectively, and $l$ represents the scale level.
% Specifically, 
we utilize level-2 Discrete Wavelet Transform (DWT) to obtain different frequency components, where $w_L$ represents the low-frequency component and $w^{(l)}_H$ represents the high-frequency components of level $l$. 
The low-frequency components $w_L$ have the smallest size ($\frac{H}{4}, \frac{W}{4}$) and are directly used to generate the lowest level of semantic feature maps $f_s^{(0)}$ via a CNN-based feature extractor. 
For each level of high-frequency components, 
it is infeasible to generate spatial features by naively adding different frequency components together due to the domain gap.
% since one cannot directly , 
We thus design an Inverse Wavelet Block (IWB) that simulates the inverse discrete wavelet transform by combining frequency features of
the previous level with high-frequency features of the current level via dilated deconvolution to generate latent spatial feature maps $f_L^{(l)}$. Then the latent spatial feature maps are used to generate semantic feature maps of the current level by CNN as below:
\begin{align}
    f_s^{(l)} = \textbf{CNN}(f_s^{(l-1)}, \textbf{IWB}(f_L^{(l-1)}, w_H^{(l)})), ~~l\in{1, 2}.
\end{align}
In addition, all the high-frequency features are eventually gathered to form the 2D compounded high-frequency components which are used to generate frequency feature maps $f_w$ by a CNN-based network. 

After having the spatial semantic feature maps and the wavelet feature maps, we follow the same approach as in CasMVSNet~\cite{gu2020cascade} to build sweep planes and spatial feature volumes $P_s^{(l)}$ at three levels.  Besides, thanks to the nice property of Wavelet Transform that it does not affect the relative coordinates, we can follow the same manner to construct the high-frequency feature volume. Since high-frequency information is often sparsely distributed, it is sufficient to represent the high-frequency features in a relatively small volume. Here we choose to use the second coarsest level $(l = 1)$ to balance the depth range and the depth sampling precision and construct a wavelet frequency feature volume $P_w$ with the size of $\frac{H}{2}\times\frac{W}{2}$. 
In a nutshell, given a set of input source views $\{I_v\}^V_{v=0}$, our WMVS module generates 2D feature maps $f_s^{(l)}, f_w$ and their corresponding 3D features volumes $P_s^{(l)}, P_w$ for subsequent modules as below:
\begin{align}
    (f_s^{(l)}, f_w, P_s^{(l)}, P_w) = \textbf{WMVS}(\{I_v\}^V_{v=0}), ~~l\in{0, 1, 2}.
\end{align}

\subsection{Frequency-guided Sampling Strategy}
\label{sec3.2}
After generating features from the WMVS module, we use the ray-casting approach to create new views. To cover the depth range, we sample $N_c$ points uniformly along each camera ray at a novel camera pose. Many previous studies ~\cite{liu2020neural, martin2021nerf, yu2021pixelnerf, wang2021ibrnet} follow the classic NeRF~\cite{mildenhall2021nerf}, sampling $N_f$ points based on the volume density distribution inferred by the $N_c$ points to approximate the object surfaces. However, this coarse-to-fine sampling strategy requires training two NeRF networks at the same time. MVSNeRF~\cite{chen2021mvsnerf} directly discards the fine sampling and claims that adding a fine sampling process cannot significantly improve the performance. 
GeoNeRF~\cite{johari2022geonerf} first estimates a set of valid coarse sample points by checking whether the coordinates lie within the valid NDC (Normalized Device Coordinate) coordination system and then randomly samples $N_f$ points around these valid coarse points. Although GeoNeRF simultaneously samples a mixture of $N_c + N_f$ points, it cannot ensure the sampled points are near the objects.

\begin{figure}[htb]
    \centering
    \includegraphics[width=1\columnwidth]{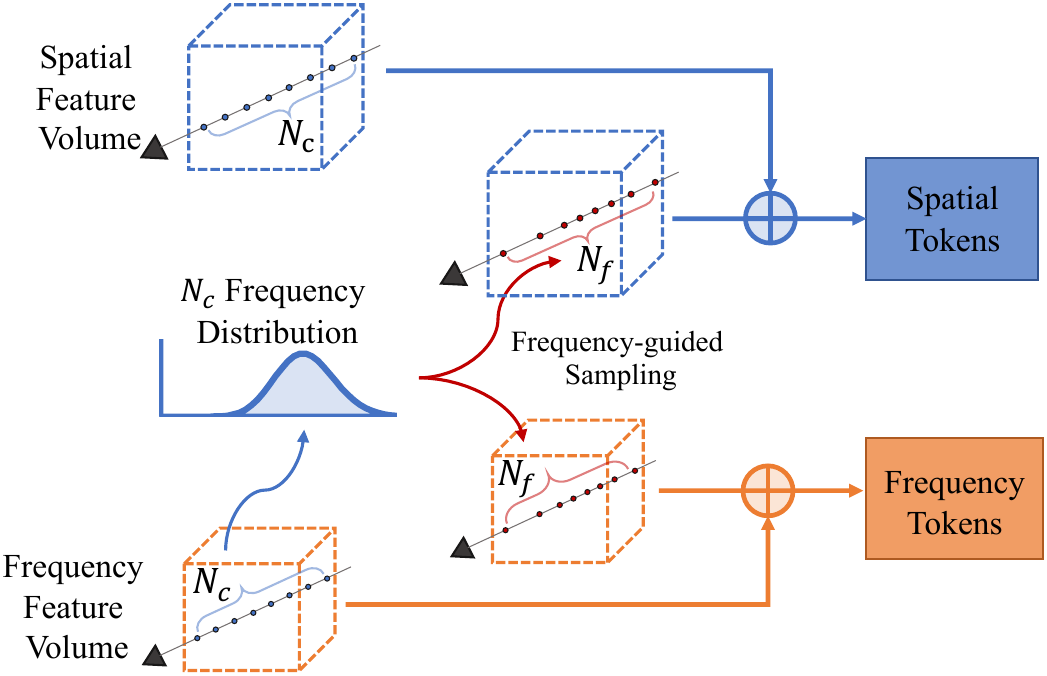}
    \caption{The illustration of our Frequency-guided Sampling Strategy (FSS). We utilize the distribution of the coarse sampling points in the frequency volume to determine the distribution of the fine sampling points. Areas having higher wavelet feature values are more likely to be sampled in the fine sampling process.}
    \label{fig:fss}
\end{figure}

We propose a frequency-guided sampling strategy (FSS) (as shown in Fig. \ref{fig:fss}) based on the observation that high-frequency features often indicate valuable scene information. Our strategy first uses the coordinates of coarse sampling points to fetch corresponding high-frequency features from the wavelet feature volume $P_w$. Then, we use these frequency features to create a probability density function $p_0$ along the ray, which determines the distribution of the fine sampling points. Regions with higher wavelet feature values have a higher probability of being sampled in the fine sampling process which yields better sampling quality.
% GeoNeRF.

\begin{figure}[htb]
    \centering
    \includegraphics[width=1\columnwidth]{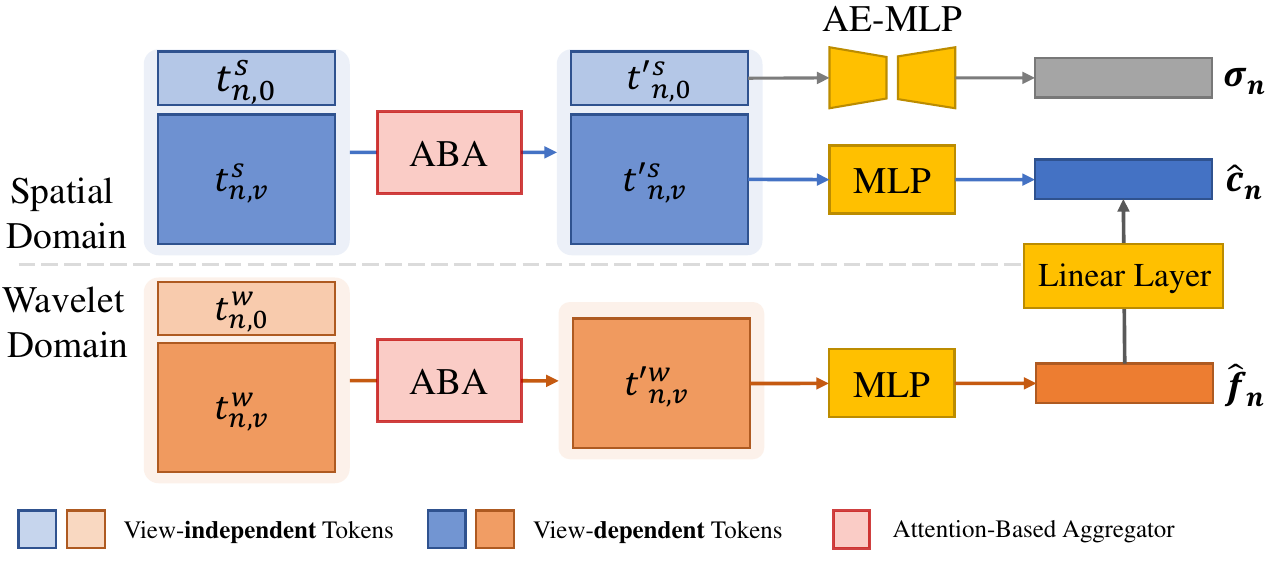}
    \caption{The overall structure of the Hybrid Neural Renderer (HNR). First, attention-based modules are employed to obtain refined tokens $\{t'^{s/w}_{n,v}\}$ for each domain. These tokens are then sent to MLP-based modules introduced in GeoNeRF~\cite{johari2022geonerf} to generate volume density $\sigma_n$, color $\hat{c}_n$, and frequency coefficient $\hat{f}_n$ for each point $x_n$. The frequency coefficient $\hat{f}_n$ is further used to adjust the color after passing through the linear layers.}
    \label{fig:hnr}
\end{figure}

\subsection{Hybrid Neural Renderer}
\label{sec3.3}

Since we have feature volumes $P_s^{(l)}, P_w$ in both the spatial domain and the frequency domain and the coordinates of the sampled points from FSS, we can fetch the features from the feature volumes and represent them as sets of tokens. For a point $x_n$ in both domains, we generate a view-independent (i.e., global) token $t_{n,0}$ and V view-dependent tokens $t_{n,v}$. We define $t^s$ and $t^w$ as tokens in the spatial domain and tokens in the wavelet frequency domain, respectively. For a sample $n$,
 $t^{s/w}_{n, 0}$ could be considered as a global understanding of the scene at point $x_n$, while $t^{s/w}_{n, v}$ represents the view-dependent understanding of the scene. We then implement a Hybrid Neural Renderer (HNR) which integrates these tokens to estimate both the colors and the frequency coefficients of the rays. The overall structure of the HNR is shown in Fig. \ref{fig:hnr}. 
 
  We first adopt an Attention-Based Aggregator(ABA) in GeoNeRF~\cite{johari2022geonerf} to refine the feature tokens. The refined view-independent tokens are used to estimate the volume density while the refined view-dependent tokens are utilized to predict the colors and frequency coefficients.
Since the global information of wavelet high-frequency is often sparse and we demand local high-frequency enhancement, we only reserve the view-independent tokens in the spatial domain for the subsequent volume density estimation. Hence, the output of ABA only contains one set of view-independent tokens $\{t'_{n, 0}\}^N_{n=1}$ which have access to all necessary data to learn the geometry of the scene and estimate volume densities. These view-independent tokens are then regularized using an auto-encoder-style MLP network (AE-MLP)~\cite{johari2022geonerf}. The AE-MLP network learns the global geometry along the ray using convolutional layers and predicts more coherent volume densities $\sigma_n$. 
 % For the view-dependent tokens, 
Notably, only the tokens in the frequency domain $\{t'^w_{n, v}\}^V_{v=1}$ are used to predict the frequency coefficients $\hat{f}_n$ while the color prediction utilizes all the view-dependent tokens. The prediction of color and frequency coefficients for each point relies on a weighted sum of the source view samples. The weight of each view, denoted as $w^{s/w}_{n, v}$, is determined using a MLP-based module. To obtain the color and wavelet samples for each point $x_n$, we project them onto the source images and the source wavelet frequency maps, resulting in the samples $c_{n,v}$ and $f_{n,v}$, respectively. We first estimate the wavelet coefficients via this weighted sum process. These wavelet coefficients form another set of weights by two linear layers which are further used to adjust the color prediction based on the weighted sum of the color samples as:
 \begin{align}
     &\hat{f_n} = \sum^V_{v=1}w^w_{n, v}f_{n,v}~, \\
     \hat{c_n} = (&\sum^V_{v=1}w^s_{n, v}c_{n,v})*(\textbf{LT}(\hat{f_n}) + 1).
 \end{align}
We argue that this design can increase the significance of the color samples around the surfaces of the objects and can reconstruct more details of the objects in the novel view. 

Once we have the prediction of the volume densities, colors, and frequency coefficients, the color and the wavelet coefficient of the camera ray at a novel pose can be estimated via the classic volume rendering technique in NeRF~\cite{mildenhall2021nerf}. Besides the color and the wavelet coefficient, we also predict the depth value of each ray for the depth supervision (see supplementary materials for more details). The volume rendering can be represented as:
\begin{align}
\begin{split}
    \{\hat{c}, \hat{f}, \hat{d}\} = \sum^N_{n=1}\text{exp}(-\sum^{n-1}_{k=1}\sigma_k)(1- \text{exp}(-\sigma_n))\{\hat{c}_n,\hat{f}_n, z_n\},
\end{split}
\end{align}
where $z_n$ is the depth of point $x_n$ with respect to the novel pose.

\definecolor{bronze}{rgb}{1,1,0.6}
\definecolor{silve}{rgb}{0.969,0.796,0.600}
\definecolor{gold}{rgb}{0.941,0.592,0.600}

\newcommand{\gold}[1]{\colorbox{gold}{{#1}}}
\newcommand{\silve}[1]{\colorbox{silve}{{#1}}}
\newcommand{\bronze}[1]{\colorbox{bronze}{{#1}}}

\begin{table*}[htb!]
\centering
\renewcommand\tabcolsep{7.5pt}
\begin{tabular}{l||ccc|ccc|ccc}
\hline
\multirow{2}{*}{Method}   & \multicolumn{3}{c|}{DTU~\cite{jensen2014large}}   & \multicolumn{3}{c|}{NeRF Synthetic~\cite{mildenhall2021nerf}}   & \multicolumn{3}{c}{LLFF~\cite{mildenhall2019local}} \\ 
\cline{2-10} & \multicolumn{1}{c|}{PSNR$\uparrow$}           & \multicolumn{1}{c|}{SSIM$\uparrow$}           & LPIPS$\downarrow$           & \multicolumn{1}{c|}{PSNR$\uparrow$}           & \multicolumn{1}{c|}{SSIM$\uparrow$}           & LPIPS$\downarrow$          & \multicolumn{1}{c|}{PSNR$\uparrow$}           & \multicolumn{1}{c|}{SSIM$\uparrow$}           & LPIPS$\downarrow$          \\ \hline
PixelNeRF~\cite{yu2021pixelnerf}                 & \multicolumn{1}{c|}{19.31}          & \multicolumn{1}{c|}{0.789}          & 0.382           & \multicolumn{1}{c|}{7.390}           & \multicolumn{1}{c|}{0.658}          & 0.411          & \multicolumn{1}{c|}{11.24}          & \multicolumn{1}{c|}{0.486}          & 0.671          \\ \hline
MVSNeRF~\cite{chen2021mvsnerf}                  & \multicolumn{1}{c|}{20.68}          & \multicolumn{1}{c|}{0.875}          & 0.243          & \multicolumn{1}{c|}{16.70}          & \multicolumn{1}{c|}{0.845}          & 0.278          & \multicolumn{1}{c|}{20.07}          & \multicolumn{1}{c|}{0.726} & 0.318          \\ \hline
PointNeRF~\cite{xu2022point}                 & \multicolumn{1}{c|}{23.89}          & \multicolumn{1}{c|}{0.874}          & 0.203           & \multicolumn{1}{c|}{22.73}              & \multicolumn{1}{c|}{0.887}              & 0.193              & \multicolumn{1}{c|}{N/A}              & \multicolumn{1}{c|}{N/A}              & N/A              \\ \hline
GeoNeRF~\cite{johari2022geonerf}      & \multicolumn{1}{c|}{\bronze{27.67}}          & \multicolumn{1}{c|}{\bronze{0.920}}          & \bronze{0.117}           & \multicolumn{1}{c|}{\bronze{24.80}}          & \multicolumn{1}{c|}{\bronze{0.891}}          & \bronze{0.182}          & \multicolumn{1}{c|}{\bronze{23.22}}          & \multicolumn{1}{c|}{\bronze{0.757}}          & \bronze{0.248}          \\ \hline
GeoNeRF* & \multicolumn{1}{c|}{\silve{29.02}}    & \multicolumn{1}{c|}{\silve{0.940}}    & \silve{0.0864}    & \multicolumn{1}{c|}{\silve{ 25.83}}    & \multicolumn{1}{c|}{\silve{0.907}}    & \silve{0.137}    & \multicolumn{1}{c|}{\gold{24.31}} & \multicolumn{1}{c|}{\silve{0.793}}          & \silve{0.213}    \\ \hline
WaveNeRF                  & \multicolumn{1}{c|}{\gold{29.55}} & \multicolumn{1}{c|}{\gold{0.948}} & \gold{0.0749} & \multicolumn{1}{c|}{\gold{26.12}} & \multicolumn{1}{c|}{\gold{0.918}} & \gold{0.113} & \multicolumn{1}{c|}{\silve{24.28}}    & \multicolumn{1}{c|}{\gold{0.794}}    & \gold{0.212} \\ \hline

\end{tabular}
\caption{Quantitative comparison of our proposed WaveNeRF with existing generalizable NeRF models in terms of PSNR$\uparrow$, SSIM$\uparrow$, and LPIPS$\downarrow$ metrics. The results in red are the best, the results in orange are the second best, and the third best ones are in yellow. 
`*' denotes that the model (i.e., GeoNeRF) was trained based on a pre-trained Cascade MVSNet checkpoint, while our model is trained from scratch. When we train the GeoNeRF model from scratch using their training scripts, its performance degrades to the values shown in the row of GeoNeRF.
}
\label{tab:overall_res}
\end{table*}

\subsection{Loss Function}
Based on previous studies, we adopt the same primary color loss $\mathcal{L}_c$ and depth loss $\mathcal{L}_D$ as GeoNeRF~\cite{johari2022geonerf}. For more details about these losses, please refer to the supplementary materials. 

In addition to these losses, we introduce two frequency losses on the predicted wavelet coefficients to supervise the training in the frequency domain. The base frequency loss function is similar to the color loss function and calculates the mean squared error between the predicted wavelet coefficients and the ground truth pixel wavelet coefficients as below:
\begin{align}
    \mathcal{L}_{f_b} = \frac{1}{|R|}\sum_{r\in R}||\hat{f}(r) - f_{gt}(r)||^2,
\end{align}
where $R$ is the set of rays in each training batch and $f_{gt}$ is the ground truth frequency coefficients.

To improve learning around high-frequency features, we have also designed a Weighted Frequency Loss (WFL), which is a modified color loss. This loss amplifies the error around the high-frequency features based on the value of the wavelet coefficients in that region. It can be represented as:
\begin{align}
    \mathcal{L}_{f_w} = \frac{1}{|R|}\sum_{r\in R}f_{gt}(r)||\hat{c}(r) - c_{gt}(r)||^2.
\end{align}
Finally, by combining all the losses mentioned above, the complete loss function of our model is represented as:
\begin{align}
    \mathcal{L} = \mathcal{L}_{c} + 0.1\mathcal{L}_{f_b} + 0.5\mathcal{L}_{f_w} + 0.1\mathcal{L}_{D}.
\end{align}

\section{Experiment}
\noindent \textbf{Dataset.}~~We have trained our generalizable network using the DTU dataset~\cite{jensen2014large}, IBRNet dataset~\cite{wang2021ibrnet}, and a real forward-facing dataset from LLFF~\cite{mildenhall2019local}. For the partition of DTU dataset, we follow the approach of PixelNeRF~\cite{yu2021pixelnerf}, resulting in 88 training scenes and 16 testing scenes while maintaining an image resolution of $600 \times 800$ as in GeoNeRF~\cite{johari2022geonerf}. For depth supervision, we only use ground truth depths from MVSNet~\cite{yao2018mvsnet} for DTU dataset. For samples from the forward-facing LLFF dataset and IBRNet dataset, we use self-supervised depth supervision. Specifically, we used 35 scenes from LLFF and 67 scenes from IBRNet as in GeoNeRF. 

To evaluate our model, we test it on three datasets: DTU test data, Synthetic NeRF data~\cite{mildenhall2021nerf}, and LLFF Forward-Facing data. DTU dataset contains 16 test scenes and the other two datasets both have 8 test scenes. We followed the same evaluation protocols as NeRF~\cite{mildenhall2021nerf} for the synthetic dataset and LLFF dataset, and the same protocol in MVSNeRF~\cite{chen2021mvsnerf} for the DTU dataset.

\noindent \textbf{Implementation details.}~~To fit the pyramid structure, we adopt a two-scale (J=2) wavelet transform for the WMVS module. Increasing the number of scales does not improve the rendering quality significantly, but it largely increases the difficulty of implementation due to the complicated padding operations. In contrast to the three different granularities ($D_s = [8, 32, 48]$) for the spatial sweep planes in WMVS, we uniformly sample 32 frequency sweep planes ($D_w = 32$) from near to far because high-frequency features are usually sparsely distributed. We set the number of points in our sampling strategy to be $N_c = 96$ and $N_f = 32$ on a ray for all scenes, and set the number of input views to be $V = 3$ for both the training and evaluation process. For more implementation details, please refer to the supplementary.

% We train our WaveNeRF model for 225k iterations using one RTX 3090 GPU. Each iteration randomly samples one scene and 512 rays are randomly selected as a training batch. We use an Adam optimizer with an initial learning rate of $5e-4$ and a cosine learning rate scheduler without restart.

\begin{figure*}[htb!htb]
    \centering
    \includegraphics[width=1.0\textwidth]{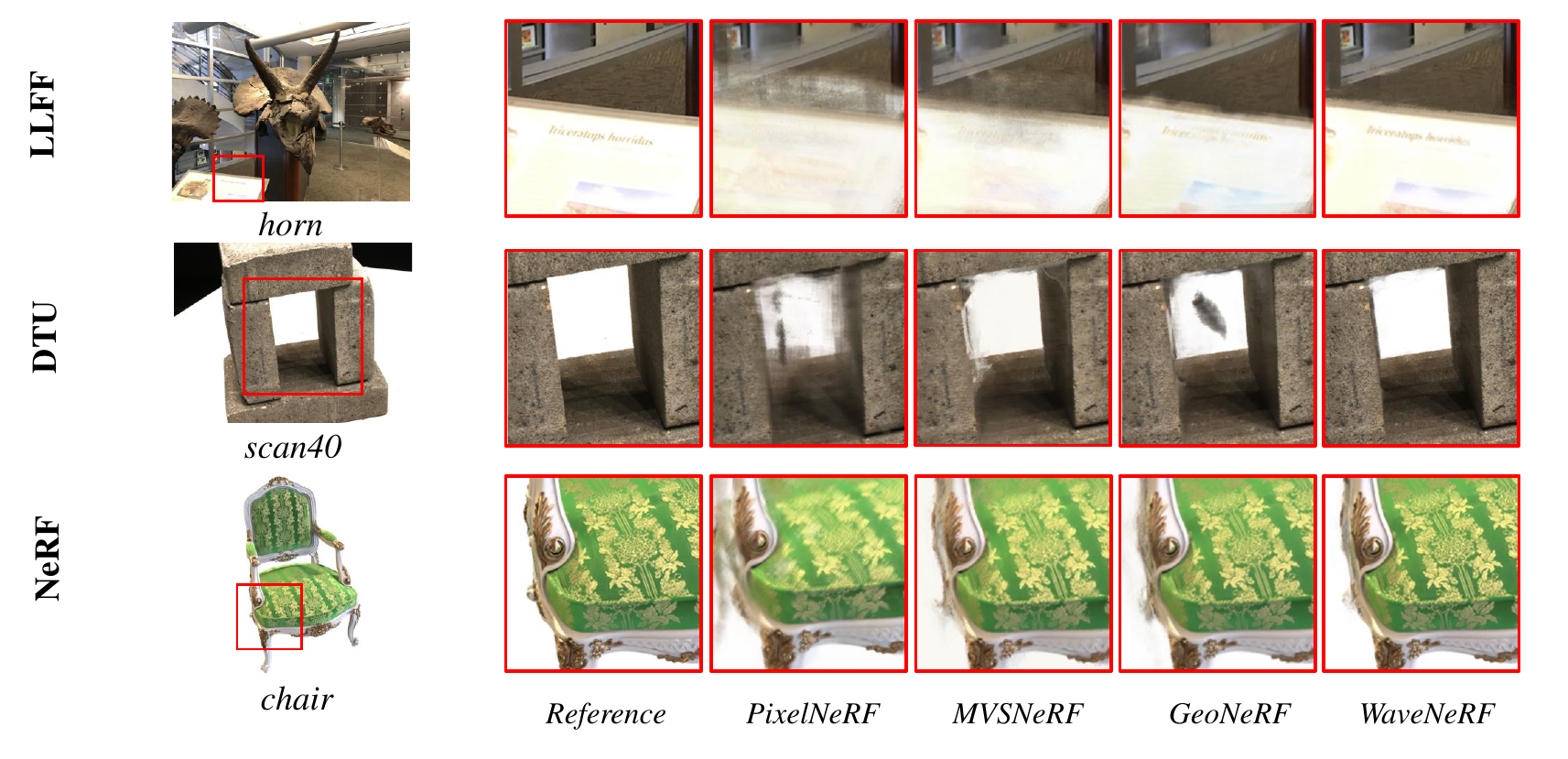}
    \caption{The qualitative results of our WaveNeRF and the comparison with PixelNeRF~\cite{yu2021pixelnerf}, MVSNet~\cite{chen2021mvsnerf}, and GeoNeRF~\cite{johari2022geonerf}. We show the scenes from LLFF dataset~\cite{mildenhall2019local} (\textit{horn}), DTU dataset~\cite{jensen2014large} (\textit{scan40}), and NeRF synthetic dataset~\cite{mildenhall2021nerf} (\textit{chair}). Our WaveNeRF model can preserve more details than the previous generalizable NeRF. }
    \label{fig:qualitative}
\end{figure*}
\begin{table*}[htb]
\centering
\resizebox{\textwidth}{!}{
\begin{tabular}{l||ccc|ccc|ccc}
\hline
\multirow{2}{*}{Experiments}                           & \multicolumn{3}{c|}{DTU~\cite{jensen2014large}}                                      & \multicolumn{3}{c|}{NeRF Synthetic~\cite{mildenhall2021nerf}}                           & \multicolumn{3}{c}{LLFF~\cite{mildenhall2019local}}                                     \\ \cline{2-10} 
                                                      & \multicolumn{1}{c|}{PSNR$\uparrow$} & \multicolumn{1}{c|}{SSIM$\uparrow$} & LPIPS$\downarrow$ & \multicolumn{1}{c|}{PSNR$\uparrow$} & \multicolumn{1}{c|}{SSIM$\uparrow$} & LPIPS$\downarrow$ & \multicolumn{1}{c|}{PSNR$\uparrow$} & \multicolumn{1}{c|}{SSIM$\uparrow$} & LPIPS$\downarrow$ \\ \hline
Baseline                                   & \multicolumn{1}{c|}{27.67}    & \multicolumn{1}{c|}{0.920}    & 0.117     & \multicolumn{1}{c|}{24.80}    & \multicolumn{1}{c|}{0.891}    & 0.182     & \multicolumn{1}{c|}{23.22}    & \multicolumn{1}{c|}{0.757}    & 0.248     \\ \hline
+ WMVS                   & \multicolumn{1}{c|}{27.97}    & \multicolumn{1}{c|}{0.922}    & 0.113     & \multicolumn{1}{c|}{24.63}    & \multicolumn{1}{c|}{0.887}    & 0.183     & \multicolumn{1}{c|}{23.23}    & \multicolumn{1}{c|}{0.762}    & 0.244     \\ \hline
+ WMVS + FSS                    & \multicolumn{1}{c|}{28.90}    & \multicolumn{1}{c|}{0.942}    & 0.084     & \multicolumn{1}{c|}{25.63}    & \multicolumn{1}{c|}{0.912}    & 0.119     & \multicolumn{1}{c|}{23.99}    & \multicolumn{1}{c|}{0.782}    & 0.227     \\ \hline
+ WMVS + FSS + HNR                        & \multicolumn{1}{c|}{29.16}    & \multicolumn{1}{c|}{0.942}    & 0.083     & \multicolumn{1}{c|}{25.89}    & \multicolumn{1}{c|}{0.916}    & 0.118     & \multicolumn{1}{c|}{24.02}    & \multicolumn{1}{c|}{0.795}    & 0.206     \\ \hline
\rowcolor{LightCyan} + WMVS + FSS + HNR + WFL                      & \multicolumn{1}{c|}{\textbf{29.55}}    & \multicolumn{1}{c|}{\textbf{0.948}}    & \textbf{0.075}     & \multicolumn{1}{c|}{\textbf{26.12}}    & \multicolumn{1}{c|}{\textbf{0.918}}    & \textbf{0.113}     & \multicolumn{1}{c|}{\textbf{24.28}}    & \multicolumn{1}{c|}{\textbf{0.794}}    & \textbf{0.212}     \\ \hline
\end{tabular}
}
\caption{The quantitative results of the Ablation studies in terms of PSNR$\uparrow$, SSIM$\uparrow$, and LPIPS$\downarrow$ metrics. The experiments are carried out on the DTU dataset, the NeRF Synthetic dataset, and the LLFF dataset. Please refer to Section \ref{sec:abla} for the details of the design of our ablation studies}
\label{tab:ablation}
\end{table*}

\subsection{Experiment Results}
We evaluate our model and compared it with existing generalizable NeRF models, including PixelNeRF~\cite{yu2021pixelnerf}, MVSNeRF~\cite{chen2021mvsnerf}, PointNeRF~\cite{xu2022point}, and GeoNeRF~\cite{johari2022geonerf}. We quantitatively compare the models in terms of PSNR, SSIM~\cite{wang2004image}, and LPIPS~\cite{zhang2018unreasonable} as shown in Table \ref{tab:overall_res}, which demonstrates the superiority of our WaveNeRF model over previous generalizable models. 
Notably, for a fair comparison, we evaluate all methods under the same setting with only three input views, and do not quote the results reported in original papers. Specifically, MVSNeRF~\cite{chen2021mvsnerf} has a nearest-view evaluation mode that uses three nearest source views for novel views, which actually imports more than three input views. 
% To restrict its input and ensure a fair comparison, 
We thus adopt its fixed-views evaluation mode that has three fixed source views.
Additionally, the pretrained checkpoints provided by GeoNeRF~\cite{johari2022geonerf} are based on the pretrained weights from CasMVSNet~\cite{gu2020cascade}, while our model is trained end-to-end. 
% To ensure fairness, 
We thus train a GeoNeRF from scratch using their scripts and evaluate both the end-to-end version and the complete version. The results show that our model can outperform GeoNeRF even if it is trained based on the pretrained weights from CasMVSNet.

In addition to quantitative comparisons, we also provide qualitative comparisons of our model with existing methods on different datasets in Fig. \ref{fig:qualitative}. Our WaveNeRF model produces images that better preserve the details of the scene and contain fewer artifacts.

\subsection{Ablation Study}
\label{sec:abla}
We conducted several ablation studies to validate the effectiveness of our designed modules on three evaluation datasets (DTU dataset~\cite{jensen2014large}, NeRF synthetic dataset~\cite{mildenhall2021nerf}, and LLFF dataset~\cite{mildenhall2019local}). The evaluation of WaveNeRF includes the following variants: 1) the baseline model without any of our novel modules, 2) the baseline model + our WMVS module, 3) the baseline model + our WMVS module + our FSS sampling strategy, 4) the baseline model + all three of our proposed modules but without the WFL loss $\mathcal{L}_{f_w}$, and 5) The complete version of our WaveNeRF model. Table \ref{tab:ablation} shows the quantitative results of the ablation study, indicating the effectiveness of our proposed modules.

\subsection{Evaluation of High-frequency Components}
To assess how well our model renders high-frequency features in images, we rely on a metric called HFIV~\cite{tan2019image}. This metric measures the proportion of high-frequency components ($\text{HF}_c$) in an image, which is indicative of its high-frequency quality. 
% Specifically, a blurry image with a higher proportion of HF components is considered to have better image quality, whereas the $HF_c$ value of a noisy image is inversely proportional to the image quality. 
% Given that our novel view reconstruction task often produces blurry images compared to the ground truth, we directly use $HF_c$ as a measure of high-frequency rendering quality. 
To facilitate comparisons across our test data, we modify HFIV to calculate the difference between the $\text{HF}_c$ of the ground truth and the $\text{HF}_c$ of the rendered results. The smaller this difference, the better the performance of the model.

We compare HFIV of our WaveNeRF, GeoNeRF~\cite{johari2022geonerf}, and MVSNeRF~\cite{chen2021mvsnerf} on the same three datasets as the previous experiments. The quantitative (see Table \ref{tab:high_freq}) results indicate that our WaveNeRF model can reconstruct better high-frequency details than the previous generalizable NeRFs.

\begin{table}[htb]
\centering
\renewcommand\tabcolsep{7.25pt}
\begin{tabular}{l|c|c|c}
\hline
Method   & DTU    & NeRF Synthetic & LLFF  \\ \hline
MVSNeRF~\cite{chen2021mvsnerf} & 0.129 & 0.1910        & 0.241 \\ \hline
GeoNeRF~\cite{johari2022geonerf}  & 0.103  & 0.0455         & 0.128 \\ \hline
\rowcolor{LightCyan} WaveNeRF & \textbf{0.0521} & \textbf{0.0362}         & \textbf{0.115} \\ \hline
\end{tabular}
\caption{Quantitative comparisons of rendered high-frequency components among MVSNeRF~\cite{chen2021mvsnerf}, GeoNeRF~\cite{johari2022geonerf}, and our WaveNeRF. The metric used here is HFIV$\downarrow$ which can measure the difference between two images on the high-frequency bands.}
% We can see from the table that our WaveNeRF model can reconstruct better high-frequency details than the previous generalizable NeRF studies.
\label{tab:high_freq}
\end{table}

\subsection{Evaluation of the Frequency-Guided Sampling}
\begin{figure}[htb]
\centering
\subfloat[LLFF]{
\includegraphics[width=0.23\textwidth]{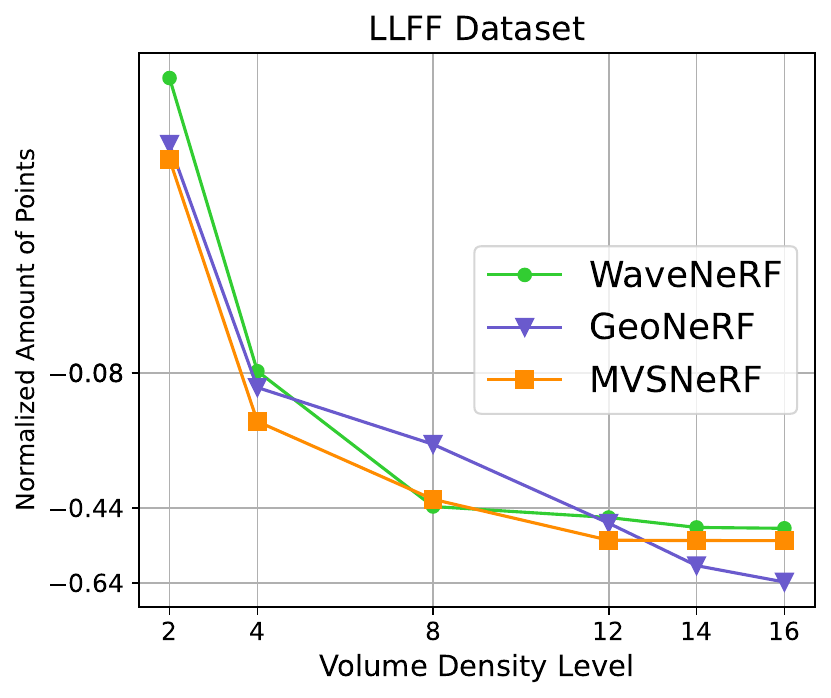}
}
\subfloat[DTU]{
\includegraphics[width=0.23\textwidth]{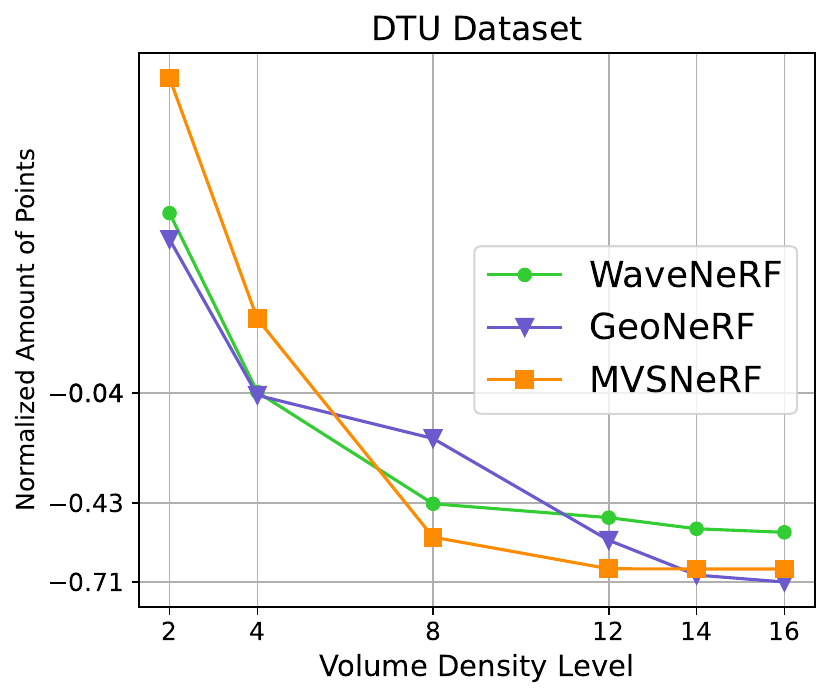}
}
\caption{Comparisons of the distribution of volume density of our WaveNeRF, GeoNeRF~\cite{johari2022geonerf}, and MVSNeRF~\cite{chen2021mvsnerf} on LLFF dataset~\cite{mildenhall2019local} and DTU dataset~\cite{jensen2014large}. The horizontal axis represents the level of volume density where large levels indicate a higher possibility of being around objects. The vertical axis means the number of sampled points whose values are standardized to a standard normal distribution for better visualization.}
% We can see from the figure that our WaveNeRF model can sample more points within the high volume density levels, which means that our FSS strategy effectively guides the model to have more samples around the surfaces of the objects.
\label{fig:dist}
\end{figure}

In the classic NeRF~\cite{mildenhall2021nerf}, the fine-sampled $N_f$ points are selected based on a normalized weight distribution obtained by estimating the volume density of coarse-sampled points, which allows to sample dense points around the region with visible content.
% samples to regions where we expect to find visible content. 
To simplify this coarse-to-fine process, GeoNeRF~\cite{johari2022geonerf} randomly samples fine points around the valid coarse points to calculate the color and the volume density of all points simultaneously. However, this randomly-sampling strategy cannot ensure that the fine-sampled points exist around the surfaces of the objects, which motivates our frequency-guided sampling strategy. In this section, we evaluate the sampling quality of our frequency-guided strategy by comparing the distribution of the volume density of the sampled points from WaveneRF, GeoNeRF~\cite{johari2022geonerf}, and MVSNeRF~\cite{chen2021mvsnerf}. As shown in Fig. \ref{fig:dist}, we can observe that our WaveNeRF model can sample more points with high volume density values, which means our FSS strategy effectively guides the model to have more samples around the surfaces of the objects.

\section{Limitation}
Our model is designed to be trained and evaluated using three-shot source views (V=3) on a single GPU with 16 GB memory. For cases with more input views, larger memory is required or the batch size should be decreased to accommodate the additional inputs.
% one will need to increase the memory available on your GPU or decrease the batch size in order to accommodate the additional inputs.
It is worth noting that our WMVS module is based on the MVS technique, which means that artifacts may appear if stereo reconstruction fails. The artifacts can manifest as noise in textureless regions or as view-dependent noisy floating-point clusters.

\section{Conclusion}
In this paper, we present a new generalizable NeRF model that is capable of generating high-quality novel view images under the few-shot setting, without requiring per-scene optimization. Our proposed model constructs MVS volumes and NeRF in the wavelet frequency domain where the explicit frequency information can be incorporated to boost the rendering quality. Additionally, we utilize frequency features to guide the sampling in NeRF, yielding densely sampled points around objects. We demonstrate that our model outperforms existing models on three datasets: the DTU dataset~\cite{jensen2014large}, the NeRF synthetic dataset~\cite{mildenhall2021nerf}, and the LLFF real forward-facing dataset~\cite{mildenhall2019local}, each with fixed-three input source views.

\section{Acknowledgements}
This work is funded by the Ministry of Education Singapore, under the Tier-2 project scheme with a project number MOE-T2EP20220-0003. Fangneng Zhan and Christian Theobalt are funded by the ERC Consolidator Grant 4DRepLy (770784).

\newpage
{\small
\bibliographystyle{ieee_fullname}
\bibliography{main}
}

\end{document}